\documentclass{article} 
\usepackage{iclr2021_conference,times}

\usepackage{amsmath,amsfonts,bm}

\def\Figref#1{Figure~\ref{#1}}

\def\Secref#1{Section~\ref{#1}}

\def\eqref#1{equation~\ref{#1}}
\def\Eqref#1{Equation~\ref{#1}}

\def\1{\bm{1}}
\newcommand{\train}{\mathcal{D}}

\def\vtheta{{\bm{\theta}}}

\def\vx{{\bm{x}}}
\def\vy{{\bm{y}}}

\def\evalpha{{\alpha}}

\DeclareMathAlphabet{\mathsfit}{\encodingdefault}{\sfdefault}{m}{sl}
\SetMathAlphabet{\mathsfit}{bold}{\encodingdefault}{\sfdefault}{bx}{n}

\def\sX{{\mathbb{X}}}
\def\sY{{\mathbb{Y}}}

\newcommand{\Ls}{\mathcal{L}}
\newcommand{\R}{\mathbb{R}}

\newcommand{\eg}{\textit{e.g.}, }
\newcommand{\ie}{\textit{i.e.}, }

\usepackage{hyperref}
\usepackage{url}

\usepackage{graphicx}

\usepackage{subcaption}

\usepackage{float}

\title{Weakly Supervised Multi-task Learning for Concept-based Explainability}

\author{Catarina Belém, Vladimir Balayan, Pedro Saleiro \& Pedro Bizarro\\
Feedzai, Lisbon, Portugal\\
\texttt{<first>.<last>@feedzai.com}\\

}

\iclrfinalcopy 
\begin{document}

\maketitle

\begin{abstract}
In ML-aided decision-making tasks, such as fraud detection or medical diagnosis, the human-in-the-loop, usually a domain-expert without technical ML knowledge, prefers high-level concept-based explanations instead of low-level explanations based on model features.
To obtain faithful concept-based explanations, we leverage multi-task learning to train a neural network that jointly learns to predict a decision task based on the predictions of a precedent explainability task (\ie multi-label concepts). There are two main challenges to overcome: concept label scarcity and the joint learning. To address both, we propose to: i) use expert rules to generate a large dataset of noisy concept labels, and ii) apply two distinct multi-task learning strategies combining noisy and golden labels. 
We compare these strategies with a fully supervised approach in a real-world fraud detection application with few golden labels available for the explainability task. With improvements of $9.26\%$ and of $417.8\%$ at the explainability and decision tasks, respectively, our results show it is possible to improve performance at both tasks by combining labels of heterogeneous quality.
\end{abstract}

%
%
%
%
%
\section{Introduction}
\label{sec:introduction}

\begin{figure}[H]
\begin{center}
\includegraphics[width=\linewidth]{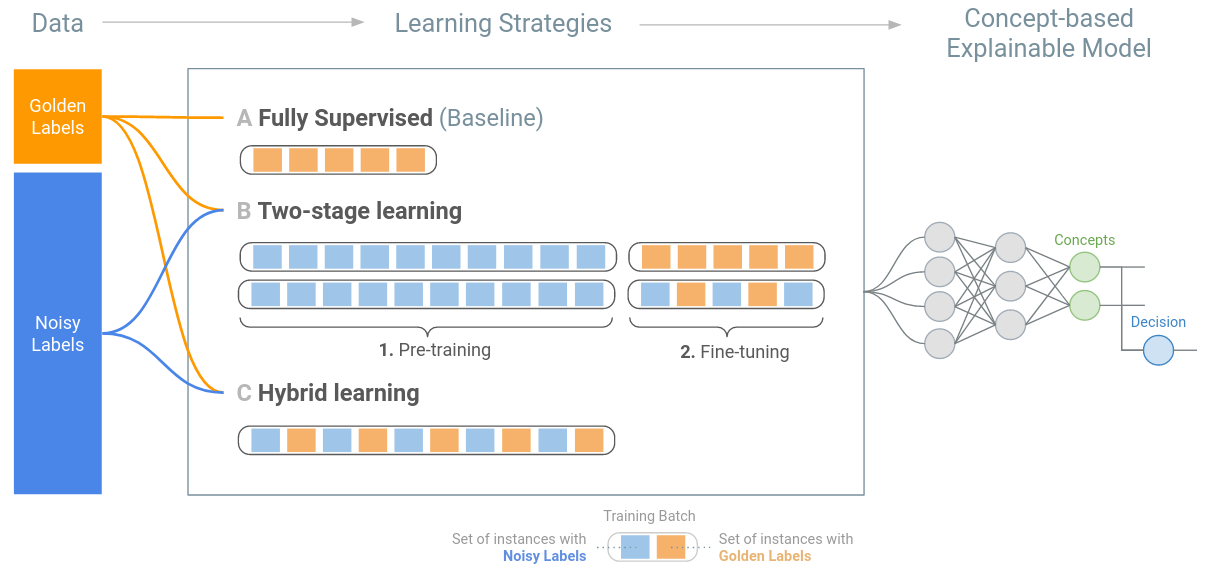} 
\end{center}
\caption{Weakly supervised multi-task learning strategies for concept-based explainability: (A) \textit{baseline} strategy resorts exclusively to golden explainability labels; (B)  \textit{two-stage learning} strategy (1) uses noisy explainability labels to pre-train a base model and (2) fine-tuning either using purely golden batches or hybrid ones; (C) \textit{hybrid learning} strategy only uses hybrid batches of golden and noisy explainability labels.}
\label{fig:distant_supervision}
\end{figure}

The AI black-box paradigm  has led to a growing demand for model explanations \citep{Ribeiro2016, Lundberg2017}. Concept-based explainability emerges as a promising family of methods addressing the information needs of humans-in-the-loop without technical ML knowledge. It concerns the generation of high-level concept-based explanations (\eg ``Suspicious payment'') rather than low-level explanations based on model features (\eg ``MCC=7801'').

Concept-based explainability can be implemented using a multi-task learning approach~\citep{Kim2018, melis2018towards,Ghorbani2019,Koh2020conceptbottleneck}. With such implementation both the decision and the explanation are learned jointly. We refer to the individual tasks as decision (or predictive) task and explainability task. Likewise, we refer to the annotation types used throughout learning as  decision (or class) labels and concepts (or explainability) labels, respectively. 

There are two main challenges towards this approach: concept label scarcity and learning to jointly predict the decision and the concepts that feed that decision. On the one hand, the creation of golden (or human) labeled datasets remains an arduous and expensive task irrespective of the application domain. 
On the other hand, the joint learning depends on several factors (\eg learning rates and/or dominance relationships of the involved tasks) and, if done incautiously, may cause deterioration of the predictions' quality.
Concept-based explainability methods must provide high-level domain knowledge explanations without compromising the quality of the conventional classification task.

This work aims to implement multi-task learning for concept-based explainability in the context of a real-world e-commerce fraud detection application. To overcome the aforementioned challenges, we first resort to weak supervision. Based on a few rule-based predictors available \textit{off-the-shelf} in historical production data, we are able to automatically generate noisy concept labels for datasets with millions of instances. Although imprecise (or weak), these noisy explainability labels prove valuable assets in training (deep) concept-based explainability models. 

Finally, since we also had access to a small set of golden explainability labels, we set out to explore learning strategies to enhance joint task performance. In particular, we explore the impact of combining different types of supervision (\ie weak and full) when training deep learning models.
\Figref{fig:distant_supervision} summarizes the learning strategies we apply.

%
%
%
%
%
\section{Preliminaries on Concept-based Explainability}
\label{sec:preliminaries}

Concept-based explainability consists of producing explanations in the format of high-level domain knowledge concepts. Following this definition, human specialists help devise a concepts taxonomy with all the relevant concepts for a specific task. These concepts closely reflect the expert's reasoning process when performing the task and therefore are perceived as suitable explanations.

Implementation-wise, this explainability paradigm can be incorporated into deep neural networks in the form of multi-task learning \citep{Ruder17a,ZhangY17Survey,melis2018towards}. To this end, the model is enlarged with an explainability (or semantic) task and the learning process is modified to allow for the joint learning of the existing decision (or predictive) task and the explainability task. 
In practice, depending on the tasks affinity, multi-task learning often boosts individual task performance \citep{Vandenhende_2021,ZhangY17Survey}.
Let $\displaystyle \train = \{ (\vx^{(i)}, \vy_D^{(i)}, \vy_E^{(i)})\}_{i=1}^{N}$ denote a dataset with $N$ instances with $d$-dimension feature vector $\displaystyle \vx \in \sX =\R^d$, $m$-dimension decision label vector $\displaystyle \vy_D \in \sY_D = \{0, 1\}^m$, and $k$-dimension explanations $\displaystyle \vy_E \in \sY_E = \{0 , 1\}^k$. The decision task is thus modeled through an $m$ multi-classification task, whereas the explainability task is modeled as a multi-label classification task with $k$ concepts.
Jointly learning the decision and explainability tasks comes down to learning the function $\displaystyle f^*: \sX \rightarrow (\sY_D, \sY_E)$.

\begin{figure}[h]
\centering
\includegraphics[width=\linewidth]{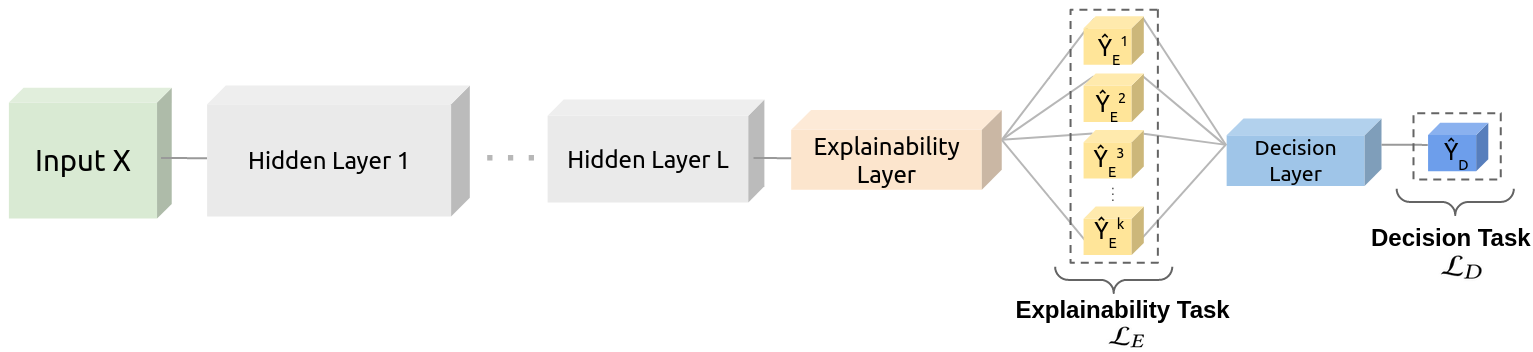}
\caption{Concept-based explainable feedforward model architecture: The explainability layer produces the concepts (yellow), which are the inputs for the decision layer. The concepts (explanations of the decision task) are also outputs of the network. Colors indicate layer type: input vector (green); hidden layer (grey); explainability layer (orange); decision layer (blue); output vectors (dashed box).}
\label{fig:conceptbasedarchitecture}
\end{figure}

\Figref{fig:conceptbasedarchitecture} shows a hard-parameter sharing approach  
towards achieving concept-based explainability~\citep{Vandenhende_2021}. In practice, we force both tasks to share the parameters of the initial layers and keep specialized output layers for each individual task.
The hierarchy observed in the output layers presupposes the explainability task carries pertinent information to the decision layer that is not explicit in the input data. 
Conversely, removing this dependency and learning both tasks in parallel may lead to learning decisions that are decoupled from the explanations.
A (deep) feedforward network with $\mathrm{L}$ hidden layers defines a mapping $\displaystyle f(\vx ; \vtheta_{1:\mathrm{L}})$, where $\displaystyle \vtheta_{1:\mathrm{L}}$ denotes all the parameters up to the $\displaystyle \mathrm{L}^{\text{th}}$ layer. 
Parameterized by $\vtheta_{E}$, the explainability layer that follows defines the mapping $\displaystyle \hat{\vy}_E = \displaystyle f_E(f(\vx ; \vtheta_{1:\mathrm{L}}); \vtheta_{E})$, hence producing a $\displaystyle k$-dimensional vector with uncalibrated probabilities for each concept. These probabilities stem from applying the sigmoid activation function (one per each neuron unit).
In addition to being explanatory, the concepts vector $\displaystyle \hat{\vy}_E$ also serves the decision layer, $\displaystyle \hat{\vy}_D = \displaystyle f_D(f_E(\vx ; \vtheta_{1:\mathrm{E}}); \vtheta_{D})$ with $\vtheta_{D}$ denoting the parameters of the decision layer. The $m$-dimension probability vector $\displaystyle \hat{\vy}_D$ results from applying the softmax function.

Learning $\displaystyle f^*$ requires mastering both the explainability and the decision tasks. 
One solution is to minimize the cross-entropy loss of each task, 
henceforth denoted $\displaystyle \Ls_E(\hat{\vy}_E, \vy_E)$ and $\displaystyle \Ls_D(\hat{\vy}_D, \vy_D)$, and combine them into a meta-loss $\displaystyle \Ls$ as defined in \Eqref{eq:training_loss}. Here, $\displaystyle \evalpha \in [0, 1]$ weighs the relative importance of the decision task over the explainability task and, thus targets different explainability-accuracy trade-offs.
\begin{equation}  
\label{eq:training_loss}  
\begin{split}
    \Ls(\vx, \vy_E, \vy_D) &= \alpha \, \Ls_D(\hat{\vy}_D, \vy_D) + (1 - \alpha) \, \Ls_E(\hat{\vy}_E, \vy_E)
\end{split}
\end{equation}

During training, the model uses backward propagation of errors (\textit{backprop}) together with mini-batch gradient descent algorithm to minimize $\displaystyle \Ls$~\citep{Rumelhart1986LearningRB}. 
Obviously,  the proposed models' generalization capacity heavily depends on the training data availability and its quality. The following section details how different approaches can be used to reach reasonable predictive performance at both tasks when departing from a small golden explainability task dataset (low-resources setting) and a large golden decision task dataset (high-resources setting).

%
%
\section{Methodology}
\label{ssec:methodology}

In this section, we start by identifying the key properties of concept-based explainability tasks that propel us into adopting a weak supervision approach. Afterwards, we propose a knowledge-based labeling technique that produces numerous but imprecise (noisy) concept labels. Finally, we put forward two learning strategies to better exploit the available labels (\ie using both noisy and golden concept labels).

\subsection{Properties for Weak Supervision}
\label{ssec:characteristics}

Despite empirical success in low-resources natural language tasks~\citep{mintz2009distant,zeng-etal-2015-distant}, weak supervision is seldom applied in algorithmic decision-making tasks. 
We identify three main characteristics, inherent to most industry AI solutions, that incentivize the use of weak supervision to make feasible the concept-based explainability paradigm.

\textbf{Abundant Golden Decision Labels.} 
Learning the mapping $\displaystyle f^*: \sX \rightarrow (\sY_D, \sY_E)$ entails having labels for both the decision and the associated concepts.
In many industry settings, it is relatively straightforward to obtain massive 
golden labeled datasets (hundreds of thousands or millions of instances) for the decision task. For example, modern financial fraud prevention systems are already designed to persist the outcome of payment transactions (legitimate or fraudulent). 

\textbf{Scarce Golden Explainability Labels.} 
Many systems are unprepared (or lack the infrastructure) for capturing specific concept annotations. 
Even in cases where companies do accrue information about the human-in-the-loop thinking process, it is frequently done in an impromptu and unstructured fashion, making it difficult and impractical to automatically process. 
Alternatively, recruiting people to hand-curate tens of thousands of instances can quickly become prohibitively time-consuming and expensive \citep{mintz2009distant} -- a cost that is further exacerbated in multi-label settings. 

\textbf{Availability of domain knowledge information.}
At the same time, modern AI-powered systems often co-exist with rule systems. For instance, in a fraud prevention solution, rules-based systems can be very effective in short-listing payment transactions based on the triggered rules. Moreover, enlarging the set of rules with additional business constraints (\eg automatically reject transactions with a specific IP) is trivial. 

\subsection{Distant Supervision}
\label{ssec:distant_supervision}

Previous research works adopt weak supervision strategies to overcome the label scarcity problem~\citep{mintz2009distant, zeng-etal-2015-distant}. In particular, they draw heuristics based on ``distant'' systems, such as databases and dictionaries, to automatically create abundant and imprecise labeled datasets. This technique is also known as \textit{distant supervision}~\citep{mintz2009distant, go2009twitter}. 

Our work applies a similar technique to overcome the concepts label scarcity inherent to concept-based explainability. We use \textit{distant supervision} to heuristically extract imprecise proxy annotations for the concepts.
We draw on the information left by rules-based systems that co-exist with real-world AI-based solutions. Through the extraction of semantic information within each rule, we build one-to-many mappings of a set of rules to the corresponding concepts in the taxonomy. Next, to prevent (some) label noise, domain experts validate the correctness and significance of these mappings.

\begin{table}[hbtp]
\caption{\textit{Rule-to-concept} mapping examples}
\label{reasons_rules_mapping}
\begin{center}
\begin{tabular}{ll}
\multicolumn{1}{c}{\bf Rule description}  &\multicolumn{1}{c}{\bf Mapped concepts}
\\ \hline \\
Order contains risky product styles.                &Suspicious Items \\
User tried \textit{n} different cards last week.    &Suspicious Customer, Suspicious Payment \\
\end{tabular}
\end{center}
\end{table}

Table~\ref{reasons_rules_mapping} presents two validated \textit{rule-to-concept} mappings in an \textit{e-commerce} fraud detection use case. Each rule comprises a human readable description with enough domain knowledge to discern the most adequate fraud concepts. A single rule may be linked with more than one domain concept in the fraud taxonomy. 
Consider a payment transaction $x \in \sX$ for which no concept labels exist and for which both rules in the table are activated. 
The proposed approach automatically attributes the labels ``Suspicious Items'' (resulting from the first rule), ``Suspicious Customer'', and ``Suspicious Payment'' (resulting from the second rule) to $x$. The true potential of this technique lies in its ability to bulk annotate large (pre-existing) data volumes, thus allowing us to quickly create multi-label datasets. 
Despite still requiring human effort to create these associations, the total human effort is negligible when compared with the manual annotation of the same volume of data.

As previously mentioned, this work pressuposes the existence of a large dataset encompassing information about the set of triggered rules as well as the decision labels (\ie the labels for the decision task $\displaystyle \sY_D$). 
Then, using the described distant supervision approach, we bulk assign noisy labels $\vy_E \in \{0, 1\}^k$. We obtain the final weakly labeled dataset, $\displaystyle \train = \{ (\vx^{(i)}, \vy_D^{(i)}, \vy_E^{(i)})\}_{i=1}^{N}$, ready to be used for model training. 
We expect this partially-weak-partially-full supervised dataset to yield significant performance gains in terms of the explainability task when compared to having no labeled data. The ensuing section focus on how to leverage the weak explainability labels obtained to improve performance through learning. 

\subsection{Learning Strategies}
\label{ssec:learning_strategies}

From an empirical standpoint, we consider the interplay between weak (or distant) supervision and multi-task learning: \textit{Are we able to outperform the fully supervised baseline}? To answer this question, we devise two learning strategies to combine the explainability labels' quality differently. With these strategies, the model is given access to larger datasets, which may benefit its generalization capabilities at both tasks.

\subsubsection{Two-stage learning strategy}

We suggest separating the learning process in two sequential stages: the \textit{pre-training stage} and the \textit{fine-tuning stage}. The former refers to the training of a base model using the large but noisy dataset, whereas the latter intends to specialize the base model using golden labels. 
Technically speaking, we use a transfer learning technique~\citep{Weiss2016transferlearning} in which we learn the model's parameters on a related dataset (the noisy dataset) and use it to obtain better performing models on the smaller target dataset (the golden-labeled dataset). In practice, we assume that, by tuning the model with a small set of examples labeled by domain-experts, it will entail improved explainability.

Notwithstanding the stated improvements on convergence and training speed during model training, if applied naively, the proposed approach can cause performance decay~\citep{Wang2018negativetransfer} -- an unpractical scenario specially in high-stakes AI applications. This is often the case when the datasets on each stage are drawn from different distributions. 
It may also occur that upon transferring this knowledge to the target dataset, the model ends up completely discarding previous information. For instance, using a learning rate value that causes steep updates or even iterating for many epochs, can be too aggressive and cause the model to unlearn the decision task. Consequently, we suggest freezing the hidden layers of the concept-based explainability model (grey layers in \Figref{fig:conceptbasedarchitecture}) and only have the task-specific layers being tweaked and learned in the \textit{fine-tuning stage}.

\subsubsection{Hybrid Learning} 

Depending on the real-world application, the explainability task is likely to assume an auxiliary role in the learning process. When training this task on the noisy labels, we risk learning a highly biased model. Although fine-tuning may help to pay off for some of the introduced bias, this strategy can still be suboptimal.

For that reason, we test a \textit{hybrid} learning strategy with the intent to promote faster and better results.
Rather than using fully distantly supervised batches to train the multi-task model, we create mixed batches with part golden, part noisy concept labels.
As a consequence, we assume that gradient updates tend to be more informative and less prone to capture noise.

%
%
%
%
%
\section{Experiments and Results}
\label{sec:experiment}

We evaluate and compare the proposed learning strategies in a real-world e-commerce fraud detection application. 
The main task, \ie the traditional decision task, aims to discern fraudulent from legitimate payment transactions (a binary classification setting). Conversely, the explainability task is perceived as an auxiliary task, whose goal is to improve the human-in-the-loop's decision-making. Using a total of $14$ domain concepts (extracted from a fraud patterns taxonomy), the explainability task concerns the attribution of the corresponding concepts to the transaction (a multi-label classification setting). 

\begin{figure}[htbp]
    \centering
    \includegraphics[width=0.6\textwidth]{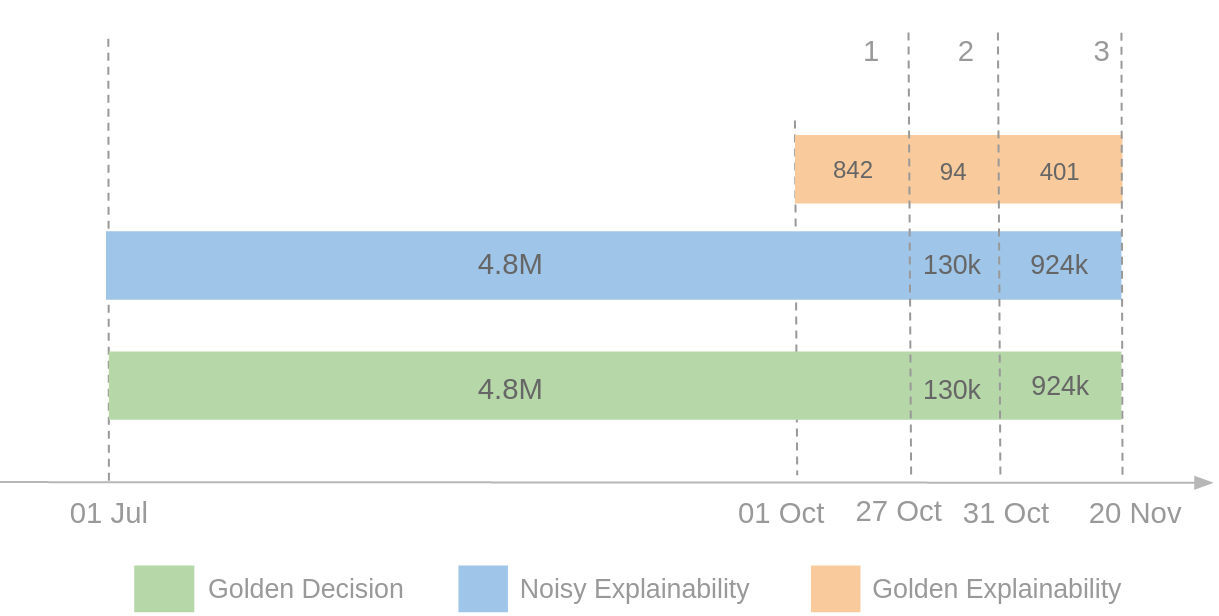}
     \caption{Datasets timeline and corresponding evaluation splits. Models are trained in the training set (1), thresholds determined in the validation set (2), and models' performance compared in the test set (3).}
    \label{fig:timeline}
\end{figure}

\textbf{Datasets.} We use a privately held dataset totalling approximately $6$ million payment transactions. Each transaction consists of information about the purchase (\eg number of items, shipping address), the fraud decision label, and the information about the triggered rules. We apply the distant supervision technique (described in \Secref{ssec:distant_supervision}) to obtain the \textit{noisy explainability} labels. 
Additionally, we have access to smaller subset of the dataset with human-annotated labels for both tasks, which totals approximately $1.3$k transactions, $37\%$ of which are fraudulent.
Note that all labels referring to the fraud decision task are golden and, henceforth, denoted as \textit{golden decision} labels. Conversely, the explainability task spans both a small \textit{golden explainability} dataset and a large \textit{noisy explainability} dataset. \Figref{fig:timeline} depicts the datasets timeline and the evaluation splits of our experiments. We follow a three-way holdout evaluation split composed of training (only $2\%$ of instances are fraudulent), validation ($4\%$ of fraud prevalence), and a test set ($4\%$ fraudulent events). These are used for training different model configurations, for selecting the decision label threshold, and for comparison between the generalization capabilities of each variant. 

\textbf{Learning Variants.} We evaluate a total of three settings (all of which depicted in \Figref{fig:distant_supervision}) using two random seeds: (1) full supervision, which trains under a fully supervised low-resources setting; (2) \textit{two-stage learning}, which first pre-trains a base model using abundant noisy explainability labels and is then refined using few golden explainability labels; and (3) \textit{hybrid learning}, which combines both noisy and golden explainability labels into the same batch during learning.

\textbf{Hyperparameter Optimization.} We run the same hyperparameter grid for each of the evaluated variants, in which we vary the number and dimension of hidden layers, learning rate, as well as the relative weight $\alpha$ of the importance task over the explainability task (see \Secref{sec:preliminaries}). A second hyperparameter grid is defined and used during the fine-tuning phase of the \textit{two-stage} learning strategy. In this grid, we vary the number of epochs, batch size, and learning rate. 

\textbf{Metrics.} We evaluate models in terms of their predictive performance at both tasks in the golden test set. For the decision task, we are restricted by business requirements to measure fraud recall at $5\%$ false positive rate (FPR), henceforth, abbreviated as recall@$5\%$. 
Conversely, we do not have any business constraint on the explainability performance metric. Instead, we use the mean Average Precision (mAP) because of its applicability and usefulness in real-world scenarios. In particular, it focus on the number of correctly predicted concepts and does not impose restrictions on the explanation size (\ie how many concepts each explanation should contain).

%
%
%
%
%

\begin{figure}[ht] 
  \begin{subfigure}[b]{0.5\linewidth}
    \centering
    \includegraphics[width=0.9\textwidth]{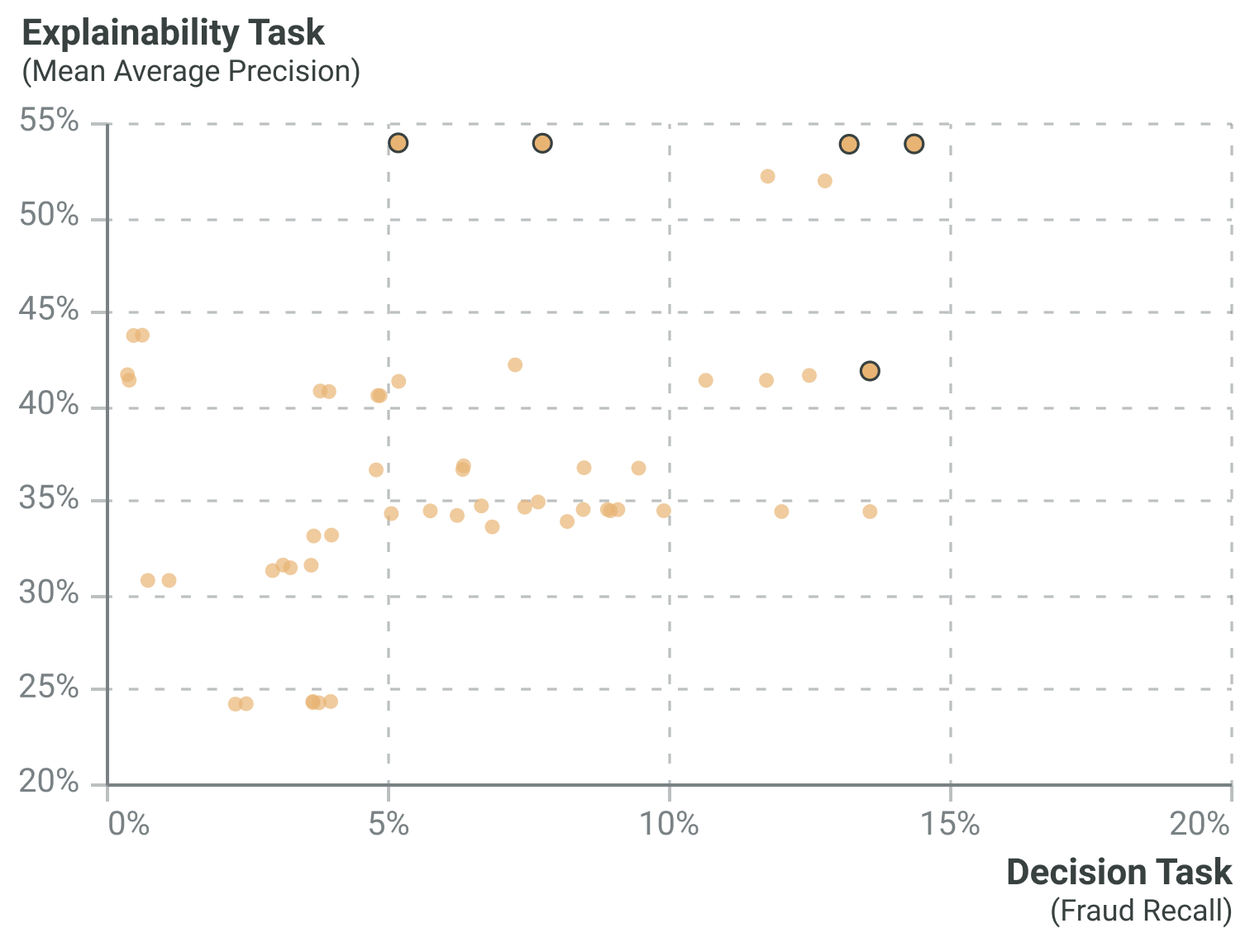}
    \caption{Fully supervised learning strategy.} 
    \label{fig:full_supervised_human} 
    \vspace{4ex}
  \end{subfigure}
  \begin{subfigure}[b]{0.5\linewidth}
    \centering
    \includegraphics[width=0.9\textwidth]{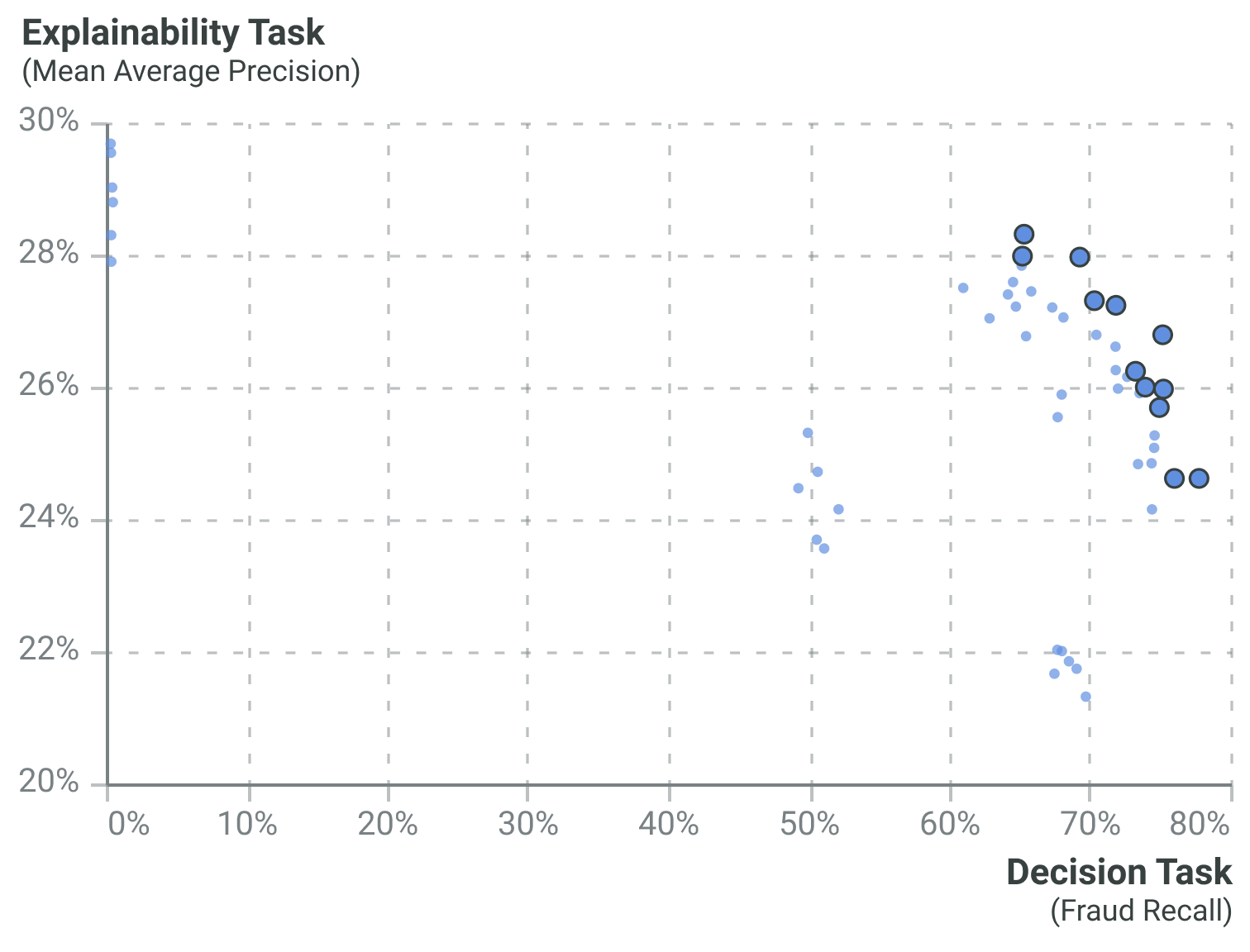}
    \caption{Two-stage learning strategy (pre-training).} 
    \label{fig:distant_supervised_base_model} 
    \vspace{4ex}
  \end{subfigure} 
  \begin{subfigure}[b]{0.5\linewidth}
    \centering
    \includegraphics[width=0.9\textwidth]{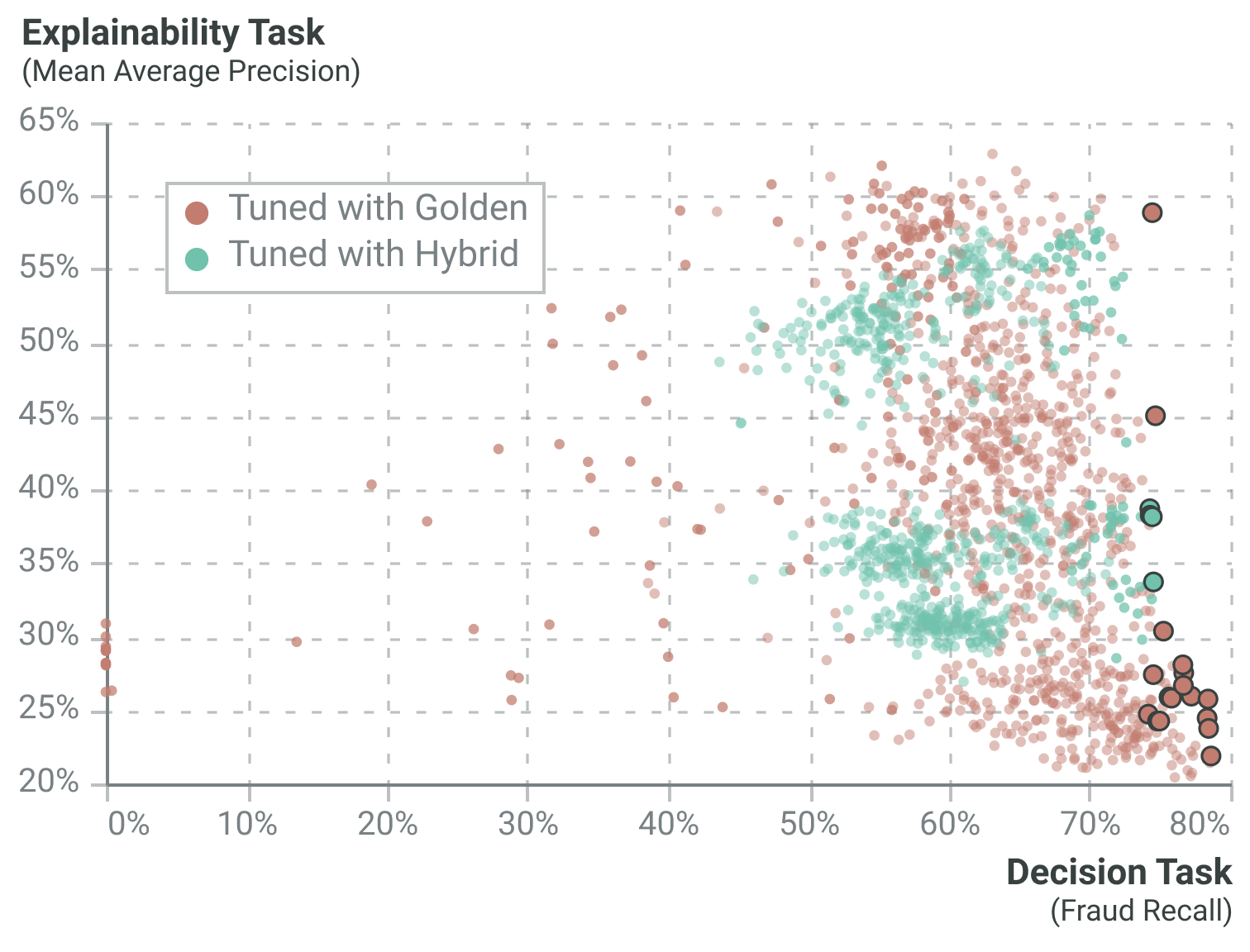}
    \caption{Two-stage strategy learning (fine-tuning).} 
    \label{fig:distant_supervised_fine_tuned} 
  \end{subfigure}
  \begin{subfigure}[b]{0.5\linewidth}
    \centering
    \includegraphics[width=0.9\textwidth]{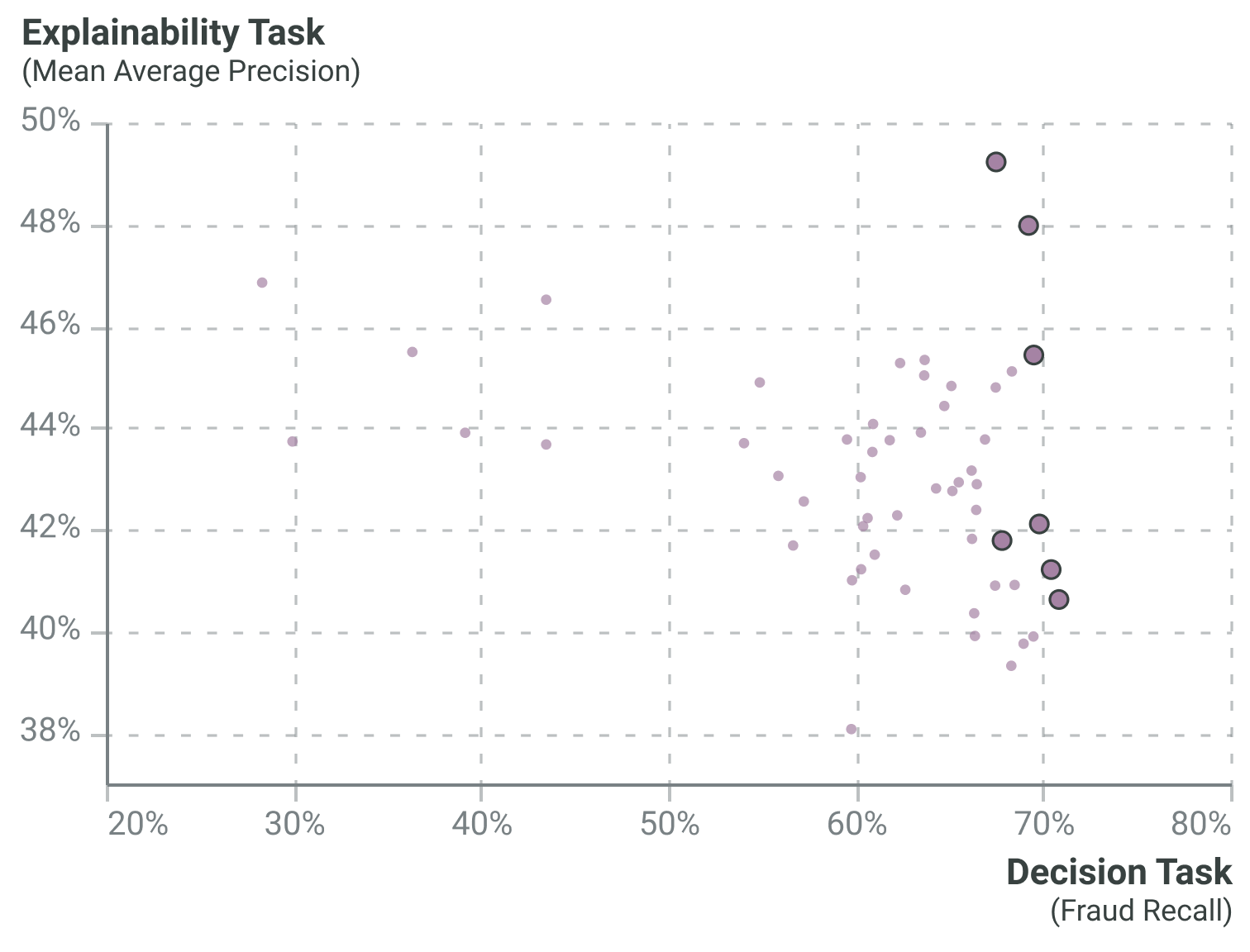}
    \caption{Hybrid learning strategy.} 
    \label{fig:hybrid} 
  \end{subfigure} 
  \caption{Explainability and predictive accuracy performance results of each learning strategy in golden test sets. These comprise the results of both random seeds. The larger-sized points represent optimal trade-offs of each learning strategy.}
  \label{fig7} 
\end{figure}

\subsection{Fully supervised Learning}
\label{ssec:baseline}
The baseline represents an aggravated low-resources setting where only a small fraction of the dataset has golden labels for both tasks and there is no other information to take advantage of. Thus, we cast the concept-based explainability multi-task problem in a supervised fashion and train models in $842$ payment transactions. In this case, we fix the fraud prevalence in the batch size at $37\%$. \Figref{fig:full_supervised_human} shows the explainability-accuracy trade-off obtained in the golden test sets for all the models obtained. The larger sized points represents the Pareto optimal models (\ie the optimal trade-offs between both tasks) at the two different runs.

Considering the performance at the decision task, most baseline models struggle to discern fraudulent from legitimate transactions with the best model achieving approximately $15\%$ recall@$5\%$. Similar results can be observed for the explainability task, though with seemingly larger mAP values. Indeed, only six models achieve mAP values larger than $50\%$. 


In general, the results seem to corroborate the idea that (deep) neural networks need massive datasets to reach peek performance. Training models for longer and more epochs in low-resource scenarios quickly results in model overfitting as observed by the higher model density in the bottom-left region of the plot. We also observe that simpler models (\ie models with fewer layers and lower learning rates) reach the best compromises in terms of recall@$5\%$ and mAP.


\subsection{Two-stage Learning}

This learning strategy is carried in two stages. The first one, dubbed the \textit{pre-training stage}, considers a high-resources scenario with approximately $4.8$ million labeled transactions. This data contains both accurate golden decision labels and imprecise concept labels (or \textit{noisy explainability} labels). Using these labels, we run the hyperparameter grid training $27$ models per random seed. From this pool of models, we pick the Pareto optimal ones, \ie the ones with the best explainability-accuracy trade-off. The selected model (dubbed base models) are then used in a second stage involving different fine-tuning configurations and a small golden explainability dataset. 

\textbf{Pre-training results.}
\Figref{fig:distant_supervised_base_model} shows the results for the first stage. We find a significant increase in the decision task performance when compared with the former fully supervised approach. As expected, all the models achieve reasonably high values of fraud recall@$5\%$ above $50\%$. This can be explained by the size of the training datasets (in the order of millions of transactions). Comparing with the baseline, this represents a boost of at least $200\%$ in fraud recall@$5\%$. Additionally, we observe that some models present very low values for decision task, while maintaining relatively ``high'' explainability. We find that these models had small learning rate and weighed more heavily the explainability task.

On the other hand, we see a tendency towards lower values at the explainability task, with mAP values falling down between $20\%$ and $30\%$. Comparing to the baseline, this performance metric deteriorates significantly. One possible reason is due to the noise associated with the explainability labels, as evidenced by the low Jaccard similarity index between both types of explainability labels.
Nonetheless, from a business perspective, the obtained trade-offs are better than the fully supervised results, since the performance at the decision task is significantly higher.

\textbf{Fine-tuning results.} For this step, we selected all the base models (as discussed in \Secref{ssec:baseline}) representing the best trade-offs (the larger sized points in the \Figref{fig:distant_supervised_base_model}). 
The second stage envisions the amelioration of base models' explainability performance through fine-tuning and different learning approaches. In particular, we experiment fine-tuning with fully supervised batches (\ie pure golden label batches), as well as with a more hybrid version (\ie involving both noisy and golden labels in the batch)

\Figref{fig:distant_supervised_fine_tuned} exhibits the results for all fine-tuned models. We can observe that both learning strategies boost the pre-trained model's performance with regards the explainability task, while maintaining similar values for decision task. However, only very few models were able to achieve better fraud recall than the base models. When comparing the performance of fine-tuned models with supervised baseline models (see \Figref{fig:full_supervised_human}), we observe not only substantial performance improvements in terms of fraud recall, but also significant increases at the explainability task. 

Interestingly, we also observe two big clusters of models trained with the hybrid fine-tuning variant (green colored dots). Despite having the same range of values for decision task with values $[0.45, 0.8$], these show two different ranges in terms of explainability performance: $[0.28, 0.4]$ and $[0.45, 0.6]$. This can be explained with the increase of the golden labels percentage in the batches, \ie the greater the golden labels' fraction in the batch, the higher the values at the explainability task.

In conclusion, comparing this strategy to the fully supervised approach (baseline), the two-stage learning strategy seems to be a strong proposal to tackle the concept-based explainability problem while using a multi-task learning approach.

\subsection{Hybrid learning} 

The last learning strategy concerns the creation of hybrid training batches that include both noisy and some pre-defined fraction of golden explainability labels. All models are trained from scratch using these hybrid batches. In this experiment, we used batches of which $10\%$ of its size consisted of golden explainability labels. \Figref{fig:hybrid} shows the obtained results. We find that most trained models achieve fraud recall values above $54\%$. When compared to the fully baseline (see \Figref{fig:full_supervised_human}), models trained with the hybrid learning strategy have a substantial improvement in decision task performance, while maintaining reasonable values for explainability. 

Although these models achieve higher explainability values when compared to the two-stage pre-trained models (see \Figref{fig:distant_supervised_base_model}), they attain lower values at the decision task (\ie lower fraud recall values). Moreover, when compared to fine-tuned models (see \Figref{fig:distant_supervised_fine_tuned}), hybrid learning models seem to perform worse at both tasks.

\subsection{Final Comparison}

\begin{figure}[h]
\centering
\includegraphics[width=0.9\linewidth]{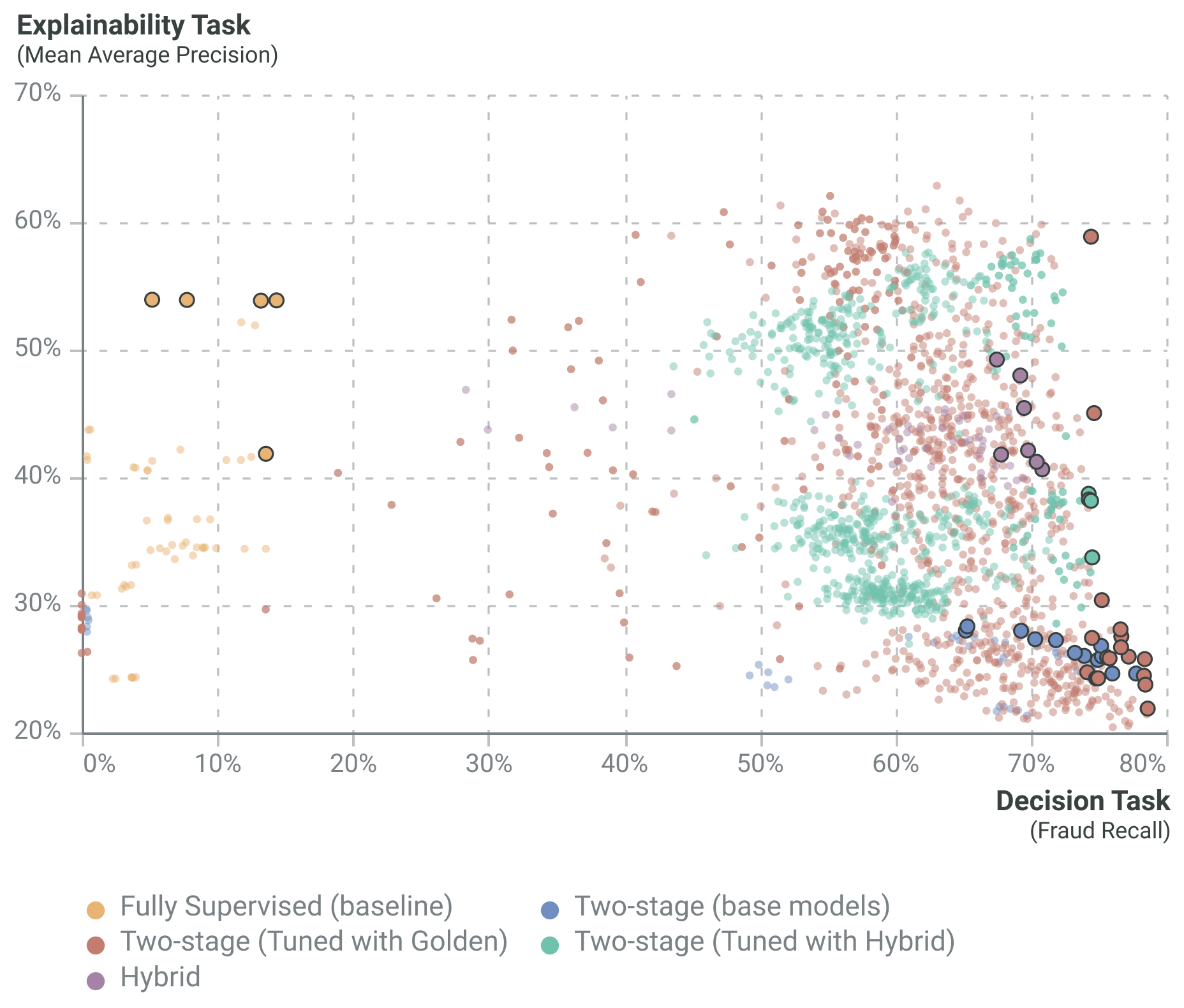}
\caption{Explainability-Accuracy comparison between the learning strategies in the test set. These comprise the results of all random seeds. The larger-sized points represent optimal trade-offs.}
\label{fig:final_comparison}
\end{figure}

In this section, we draw a comparison between each learning strategy in the test set. The results are shown in \Figref{fig:final_comparison}. These seem to corroborate the idea that it is indeed possible to jointly learn a predictive (decision) task and the associated explanations. Moreover, when compared with the low-resources fully supervision learning strategy, both proposed strategies seem to lead to significant improvements in terms of decision performance at a reduced cost in the explainability performance. In fact, in the case of the two-stage learning strategy we observe that it is possible to improve performance at both tasks simultaneously.

The boost in the decision task over the baseline is of no surprise, since the baseline is restricted to use a very small dataset (with $842$ transactions) with golden concept (or explainability) labels for both tasks. As a result, the model is not able to generalize well for unforeseen instances in the test set, thus presenting lower decision performance. On the other hand, when using weakly-labeled concepts we are able to bulk annotate massive golden decision datasets with the corresponding noisy explainability labels. Having a larger pool of golden labels, these models are able to generalize better.

Considering the hybrid learning strategy, this seems to yield consistently good trade-offs between explainability and decision tasks. On the other hand, the two-stage learning strategy spans a wider region of the explainability-accuracy solution space, reaching the best trade-offs in terms of decision task, as well as at the explainability task. Interestingly, we observe that the best trade-off\footnote{Due to business constraints, the best trade-off is the one that does not hurt performance too much and that attains reasonably high explainability performance.} is achieved by a model, trained with the two-stage learning strategy and tuned with pure golden batches. This model (top-right larger-sized orange point) achieves considerably high values in explainability, while mantaining high values at the decision task.

Finally, these results show that it is possible to learn more efficiently in a low-resources settings when using weak supervision to produce a higher-resources and noisier dataset. Moreover, depending on the importance of the two tasks, we conclude empirically that both proposed learning strategies can be used in practice with satisfactory results. Both two-stage and hybrid learning strategies improves significantly the performance on decision task at reduced (and so times at no) cost in explainability performance. We also argue that there is no one-size-fits-all learning strategy and that further experiments comparing the two proposed learning strategies are required (\eg use more random seeds, explore a wider region of the models' hyperparameter space, explore different golden label fractions for the hybrid strategy).

\section{Conclusions}
\label{sec:conclusions}

Concept-based explainability is particularly useful to explain the predictions of a black-box ML model to a non-technical, but domain expert, human-in-the-loop. A natural approach consists of training a (deep) neural network to jointly learn the predictions of a decision task and associated concepts. However, this approach faces two main challenges: concept label scarcity and the joint learning itself. We proposed to overcome these issues through the use of weak supervision approach that leverages available off-the-shelf information about expert rules to generate noisy concept labels. Furthermore, having access to a small set of golden concept labels, we devise two learning strategies to better exploit the different concept label signals (noisy and golden) during training. When comparing with the low-resources fully supervised approach, obtained results (in a e-commerce fraud detection use case) show it is possible to improve decision task performance with no (or very reduced) costs in the explainability task performance.     

\subsubsection*{Acknowledgments}
The project CAMELOT (reference POCI-01-0247-FEDER-045915) leading to this work is co-financed by the ERDF - European Regional Development Fund through the Operational Program for Competitiveness and Internationalisation - COMPETE 2020, the North Portugal Regional Operational Program - NORTE 2020 and by the Portuguese Foundation for Science and Technology - FCT under the CMU Portugal international partnership. The authors would also like to thank Beatriz Malveiro and João Palmeiro for their help with graphics editing.

\bibliography{explainable_AI,feedzai,multitasklearning,noisy_labels}

\begin{thebibliography}{15}
\providecommand{\natexlab}[1]{#1}
\providecommand{\url}[1]{\texttt{#1}}
\expandafter\ifx\csname urlstyle\endcsname\relax
  \providecommand{\doi}[1]{doi: #1}\else
  \providecommand{\doi}{doi: \begingroup \urlstyle{rm}\Url}\fi

\bibitem[Ghorbani et~al.(2019)Ghorbani, Wexler, Zou, and Kim]{Ghorbani2019}
Amirata Ghorbani, James Wexler, James~Y Zou, and Been Kim.
\newblock Towards automatic concept-based explanations.
\newblock In \emph{Advances in Neural Information Processing Systems}, pp.\
  9277--9286, 2019.

\bibitem[Go et~al.(2009)Go, Bhayani, and Huang]{go2009twitter}
Alec Go, Richa Bhayani, and Lei Huang.
\newblock Twitter sentiment classification using distant supervision.
\newblock \emph{CS224N project report, Stanford}, 1\penalty0 (12):\penalty0
  2009, 2009.

\bibitem[Kim et~al.(2018)Kim, Wattenberg, Gilmer, Cai, Wexler, Viegas, and
  Sayres]{Kim2018}
Been Kim, Martin Wattenberg, Justin Gilmer, Carrie Cai, James Wexler, Fernanda
  Viegas, and Rory Sayres.
\newblock {Interpretability beyond feature attribution: Quantitative Testing
  with Concept Activation Vectors (TCAV)}.
\newblock \emph{35th International Conference on Machine Learning, ICML 2018},
  6:\penalty0 4186--4195, 2018.

\bibitem[Koh et~al.(2020)Koh, Nguyen, Tang, Mussmann, Pierson, Kim, and
  Liang]{Koh2020conceptbottleneck}
Pang~Wei Koh, Thao Nguyen, Yew~Siang Tang, Stephen Mussmann, Emma Pierson, Been
  Kim, and Percy Liang.
\newblock Concept bottleneck models.
\newblock In Hal~Daumé III and Aarti Singh (eds.), \emph{Proceedings of the
  37th International Conference on Machine Learning}, volume 119 of
  \emph{Proceedings of Machine Learning Research}, pp.\  5338--5348. PMLR,
  13--18 Jul 2020.
\newblock URL \url{http://proceedings.mlr.press/v119/koh20a.html}.

\bibitem[Lundberg \& Lee(2017)Lundberg and Lee]{Lundberg2017}
Scott~M. Lundberg and Su~In Lee.
\newblock {A unified approach to interpreting model predictions}.
\newblock \emph{Advances in Neural Information Processing Systems},
  2017-Decem\penalty0 (Section 2):\penalty0 4766--4775, 2017.
\newblock ISSN 10495258.

\bibitem[Melis \& Jaakkola(2018)Melis and Jaakkola]{melis2018towards}
David~Alvarez Melis and Tommi Jaakkola.
\newblock Towards robust interpretability with self-explaining neural networks.
\newblock In \emph{Advances in Neural Information Processing Systems}, pp.\
  7775--7784, 2018.

\bibitem[Mintz et~al.(2009)Mintz, Bills, Snow, and Jurafsky]{mintz2009distant}
Mike Mintz, Steven Bills, Rion Snow, and Dan Jurafsky.
\newblock Distant supervision for relation extraction without labeled data.
\newblock In \emph{Proceedings of the Joint Conference of the 47th Annual
  Meeting of the ACL and the 4th International Joint Conference on Natural
  Language Processing of the AFNLP}, pp.\  1003--1011, 2009.

\bibitem[Ribeiro et~al.(2016)Ribeiro, Singh, and Guestrin]{Ribeiro2016}
Marco~Tulio Ribeiro, Sameer Singh, and Carlos Guestrin.
\newblock {"Why should I trust you?" Explaining the predictions of any
  classifier}.
\newblock \emph{Proceedings of the ACM SIGKDD International Conference on
  Knowledge Discovery and Data Mining}, 13-17-August-2016:\penalty0 1135--1144,
  2016.
\newblock \doi{10.1145/2939672.2939778}.

\bibitem[Ruder(2017)]{Ruder17a}
Sebastian Ruder.
\newblock An overview of multi-task learning in deep neural networks.
\newblock \emph{CoRR}, abs/1706.05098, 2017.
\newblock URL \url{http://arxiv.org/abs/1706.05098}.

\bibitem[Rumelhart et~al.(1986)Rumelhart, Hinton, and
  Williams]{Rumelhart1986LearningRB}
David~E. Rumelhart, Geoffrey~E. Hinton, and Ronald~J. Williams.
\newblock Learning representations by back-propagating errors.
\newblock \emph{Nature}, 323:\penalty0 533--536, 1986.

\bibitem[Vandenhende et~al.(2021)Vandenhende, Georgoulis, Van~Gansbeke,
  Proesmans, Dai, and Van~Gool]{Vandenhende_2021}
Simon Vandenhende, Stamatios Georgoulis, Wouter Van~Gansbeke, Marc Proesmans,
  Dengxin Dai, and Luc Van~Gool.
\newblock Multi-task learning for dense prediction tasks: A survey.
\newblock \emph{IEEE Transactions on Pattern Analysis and Machine
  Intelligence}, pp.\  1–1, 2021.
\newblock ISSN 1939-3539.
\newblock \doi{10.1109/tpami.2021.3054719}.
\newblock URL \url{http://dx.doi.org/10.1109/TPAMI.2021.3054719}.

\bibitem[Wang et~al.(2018)Wang, Dai, P{\'{o}}czos, and
  Carbonell]{Wang2018negativetransfer}
Zirui Wang, Zihang Dai, Barnab{\'{a}}s P{\'{o}}czos, and Jaime~G. Carbonell.
\newblock Characterizing and avoiding negative transfer.
\newblock \emph{CoRR}, abs/1811.09751, 2018.
\newblock URL \url{http://arxiv.org/abs/1811.09751}.

\bibitem[Weiss et~al.(2016)Weiss, Khoshgoftaar, and
  Wang]{Weiss2016transferlearning}
Karl Weiss, Taghi Khoshgoftaar, and DingDing Wang.
\newblock A survey of transfer learning.
\newblock \emph{Journal of Big Data}, 3, 05 2016.
\newblock \doi{10.1186/s40537-016-0043-6}.

\bibitem[Zeng et~al.(2015)Zeng, Liu, Chen, and Zhao]{zeng-etal-2015-distant}
Daojian Zeng, Kang Liu, Yubo Chen, and Jun Zhao.
\newblock Distant supervision for relation extraction via piecewise
  convolutional neural networks.
\newblock In \emph{Proceedings of the 2015 Conference on Empirical Methods in
  Natural Language Processing}, pp.\  1753--1762, Lisbon, Portugal, September
  2015. Association for Computational Linguistics.
\newblock \doi{10.18653/v1/D15-1203}.
\newblock URL \url{https://www.aclweb.org/anthology/D15-1203}.

\bibitem[Zhang \& Yang(2017)Zhang and Yang]{ZhangY17Survey}
Yu~Zhang and Qiang Yang.
\newblock A survey on multi-task learning.
\newblock \emph{CoRR}, abs/1707.08114, 2017.
\newblock URL \url{http://arxiv.org/abs/1707.08114}.

\end{thebibliography}
\bibliographystyle{iclr2021_conference}

\appendix

\end{document}